\documentclass[sigconf, nonacm, screen]{acmart}

\AtBeginDocument{%
  \providecommand\BibTeX{{%
    \normalfont B\kern-0.5em{\scshape i\kern-0.25em b}\kern-0.8em\TeX}}}

\copyrightyear{2022}
\acmYear{2022}
\setcopyright{acmcopyright}\acmConference[WWW '22 Companion]{Companion Proceedings of the Web Conference 2022}{April 25--29, 2022}{Virtual Event, Lyon, France}
\acmBooktitle{Companion Proceedings of the Web Conference 2022 (WWW '22 Companion), April 25--29, 2022, Virtual Event, Lyon, France}
\acmPrice{15.00}
\acmDOI{10.1145/3487553.3524649}
\acmISBN{978-1-4503-9130-6/22/04}

\usepackage[whole]{bxcjkjatype}
\usepackage{amsmath}
\usepackage{bm}  %
\usepackage{amsfonts}
\usepackage{booktabs}

\begin{document}

\title{ViNTER: Image Narrative Generation \\with Emotion-Arc-Aware Transformer}
\renewcommand{\shorttitle}{ViNTER: Image Narrative Generation with Emotion-Arc-Aware Transformer}

\author{Kohei Uehara}
\authornote{Both authors contributed equally to this research.}
\email{uehara@mi.t.u-tokyo.ac.jp}
\affiliation{%
  \institution{The University of Tokyo}
  \city{Tokyo}
  \country{Japan}
}

\author{Yusuke Mori}
\email{mori@mi.t.u-tokyo.ac.jp}
\authornotemark[1]
\affiliation{%
  \institution{The University of Tokyo}
  \city{Tokyo}
  \country{Japan}
}

\author{Yusuke Mukuta}
\email{mukuta@mi.t.u-tokyo.ac.jp}
\affiliation{%
  \institution{The University of Tokyo / RIKEN}
  \city{Tokyo}
  \country{Japan}
}

\author{Tatsuya Harada}
\email{harada@mi.t.u-tokyo.ac.jp}
\affiliation{%
  \institution{The University of Tokyo / RIKEN}
  \city{Tokyo}
  \country{Japan}
}

\renewcommand{\shortauthors}{Kohei Uehara, Yusuke Mori, Yusuke Mukuta, and Tatsuya Harada}

\begin{abstract}
Image narrative generation is a task to create a story from an image with a subjective viewpoint.
Given the importance of the subjective feelings of writers, readers, and characters in storytelling, an image narrative generation method should consider human emotion.
In this study, we propose a novel method of image narrative generation called ViNTER (Visual Narrative Transformer with Emotion arc Representation), which takes ``emotion arc'' as input to capture a sequence of emotional changes.
Since emotion arcs represent the trajectory of emotional change, it is expected that we can include detailed information about the emotional changes in the story to the model.
We present experimental results of both automatic and manual evaluations on the Image Narrative dataset and demonstrate the effectiveness of the proposed approach.  
\end{abstract}



\maketitle

\section{Introduction}

\begin{figure}[!t]
\centering
\includegraphics[width=0.8\columnwidth]{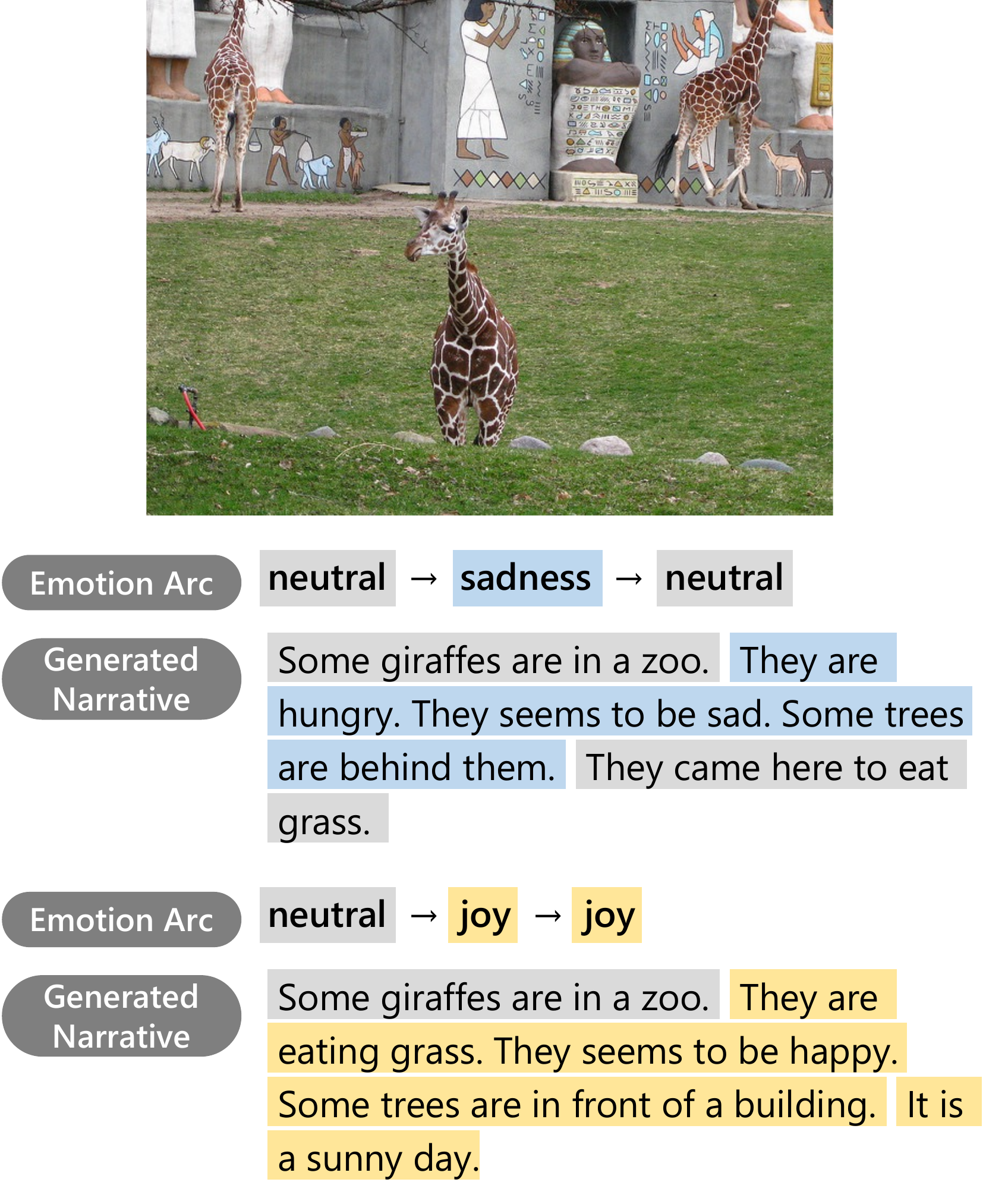}
\caption{
ViNTER is able to control the emotional changes in the generated narratives by utilizing ``emotion arc.''
For example, if the input emotion arc is ``neutral $\rightarrow$ sadness $\rightarrow$ neutral'', the model produces a narrative in which the middle part represents a sad feeling.
In contrast, if we input ``neutral $\rightarrow$ joy $\rightarrow$ joy'' as the emotion arc, the model produces a narrative filled with happy emotions from the midpoint onward.
}
\label{fig:generation_example}
\end{figure}
Image captioning is a representative and essential task in research on vision and language.
Beyond the generation of accurate descriptive caption about images, recent research efforts have focused on creative text generation, such as narrative generation from a single image~\cite{shin2018customized}, narrative generation from sequences of images~\cite{huang-etal-2016-visual,liu2017let}, and poetry generation from an image~\cite{liu2018beyond}.
These methods, which are designed to generate creative text from an image or sequence of images, can be considered as an important step toward creative artificial intelligence (creative AI).

In this paper, the term \textbf{image narrative} is used according to its definition in \cite{shin2018customized}.
That is, an image narrative is characterized by the presence of a broader range of subjective and inferential elements, unlike descriptive image caption, which describe factual aspects of the visual contents in the image.

In this paper, we focus on ``emotion'' as an essential element of image narratives.
The relationship between narratives and emotions has been an essential area of research in the humanities, represented by scientific investigations into the cognitive and effective impacts of literature~\cite{Hogan_2006_10.2307/25115327,Pandit2006,Johnson-Laird_2008-07784-007,Hogan_2010_10.5250/symploke.18.1-2.0065,Hogan2019}.
These findings have been introduced in natural language processing (NLP) field~\cite{Anderson_and_McMaster_1982} and 
various studies have been conducted on understanding and generating narratives, such as story generation with sentiment control~\cite{luo-etal-2019-learning,Dathathri2020Plug,MEGATRON-CNTRL}.

In this study, we propose a novel image narrative generation model called ViNTER (Visual Narrative Transformer with Emotion Arc Representation), which is designed to generate a narrative from an image using Transformer-based pre-trained models.
To give the model the ability of controlling emotional change while generating narratives, we utilize ``emotion arc,'' which represents the trajectory of emotion changes~\cite{EmotionalArcs}.

Emotion arc is denoted as a sequence of three words that describe the emotion, e.g., ``joy $\rightarrow$ fear $\rightarrow$ surprise.''
Each emotion word represents the emotion of the begin, body, and end parts of the narrative, respectively.
In this study, an image narrative consists of five sentences, so the begin part is defined as the first sentence, the body part as the next three sentences, and the end part as the last sentence.

For example, in the example in Figure~\ref{fig:generation_example}, the input for the emotion arc is ``neutral $\rightarrow$ sadness $\rightarrow$ neutral.''
Hence, the model is expected to generate narratives in which the body part represents a sad emotion (e.g., ``They are hungry. They seems to be sad. Some trees are behind them'').
In contrast, if the input emotion arc is ``neutral $\rightarrow$ joy $\rightarrow$ joy'', the model is expected to generate narratives with happy emotions after the middle part.

Inspired by recent studies in multi-modal Transformer models~\cite{chen2020uniter}, we designed an encoder that takes multi-modal input, an image and emotion arc.
Here, emotion arc is considered as the sequence of emotion words.
Then, a Transformer-based decoder generates narratives from the fused feature obtained from the encoder.

We evaluated our model with both of the automatic and manual evaluation.
The results show that our model demonstrates sufficiently promising performance for generating narratives that take into account emotion changes.

Our main contributions are summarized as follows:

\begin{itemize}
    \item We propose ViNTER, a novel image-narrative generation method.
    Benefiting from recent advances in multi-modal Transformer-based pre-trained models and emotion-awareness, ViNTER exhibited notable performance on image-narrative generation tasks. 
    \item  We discuss the applicability of Transformer-based models to creative AI in the fields of computer vision and language processing.
\end{itemize}

\section{Related Work}
\label{sec:related_work}

\begin{figure*}[!t]
\centering
\includegraphics[width=1.0\linewidth]{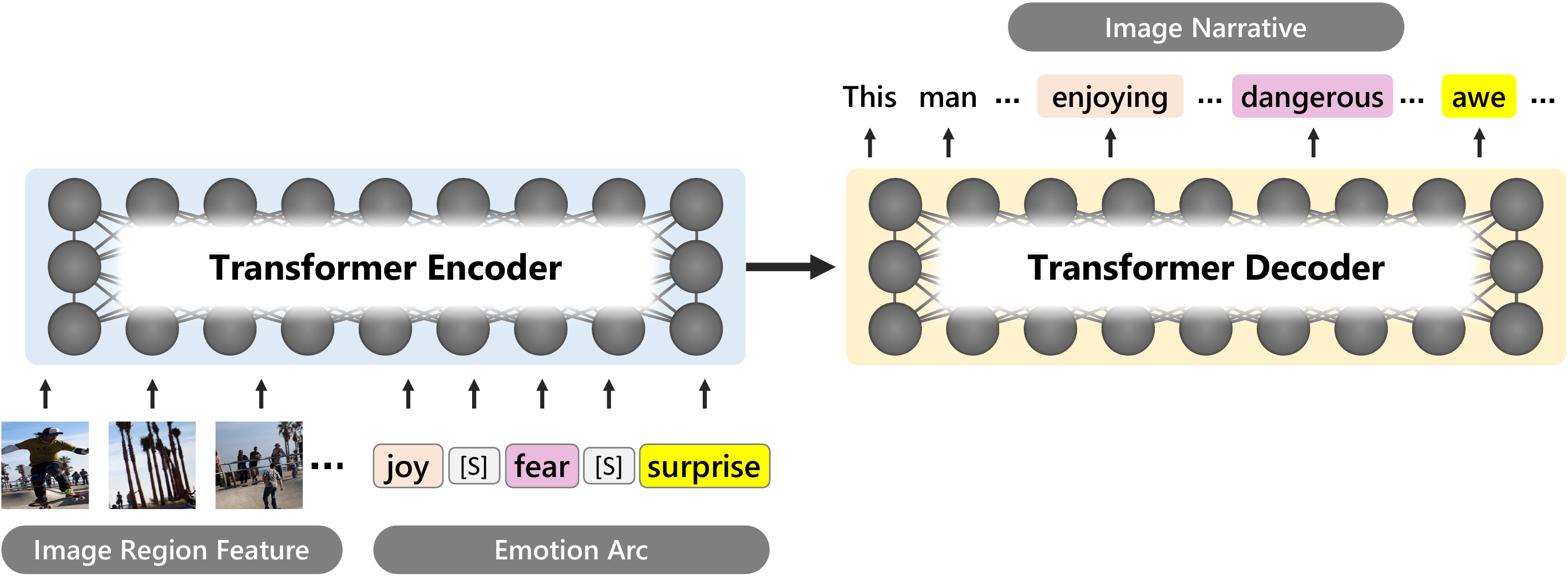}
\caption{
Overview of ViNTER.
First, we extract region-based features from an image and tokenize emotion arc into a sequence of emotion word tokens.
Then, they are fed into the bi-directional Transformer encoder.
The output feature from the encoder is fed into the auto-regressive Transformer decoder and the decoder generates each word of the narrative sequentially.
}
\label{fig:model}
\end{figure*}

\subsection{Emotion in Narratives}
\label{subsec:emotion_in_narratives}

Emotion is an essential element in storytelling. Referring to~\citet{angela_and_becca_emotion_thesaurus}, \citet{kim-klinger-2019-analysis} observed that emotion is a key component of every character, and analyzed how emotions were expressed non-verbally in a corpus of short fanfiction  stories.
\citet{Lugmayr_2017_serious_storytelling} also noted emotions as a fundamental aspect of storytelling as part of the cognitive aspects evoked by a story in its audience.

As described by~\citet{Anderson_and_McMaster_1982}, numerous efforts have been made to elucidate the relationship between emotions and stories.
Based on the 1,000 most frequently used English words for which~\citet{Heise1965} obtained semantic differential scores,~\citet{Anderson_and_McMaster_1982} reported the development of a computer program designed to assist in the analysis and modeling of emotional tone in text by identifying those words in passages of discourse.
Emotional analysis of text is closely related to lexicons annotated with emotions, and psychological findings have been referred to in these annotations and analyses.

\citet{Kim2019b} pointed out three theories of emotion commonly used in computational analyses, including Ekman's theory of basic emotions~\cite{Ekman1993}, Plutchik's wheel of emotion~\cite{plutchik1980emotion}, and Russell's circumplex model~\cite{Russell1980}.

Several studies have attempted to control story generation by considering emotions~\cite{chandu-etal-2019-way,luo-etal-2019-learning,brahman-chaturvedi-2020-modeling,Dathathri2020Plug,MEGATRON-CNTRL}.
Most of these works have focused on sentiment control, that is, positive or negative control of story generation.
Story generation with a greater variety of emotions had not been addressed until recently.
To the best of our knowledge,~\citet{brahman-chaturvedi-2020-modeling} performed the first work on emotion-aware storytelling, which considered the ``emotion arc'' of a protagonist.

Referring to a talk by Kurt Vonnegut~\cite{Vonnegut1981_video}, a famous American writer, attempts have been made to classify stories by drawing an ``emotion arc''~\cite{EmotionalArcs,Chu_and_Roy_2017_ICDM,somasundaran-etal-2020-emotion,Vecchio_2020_Hollywood_emotional_arc}.
\citet{EmotionalArcs} showed that stories collected from Project Gutenberg\footnote{http://www.gutenberg.org/} could be classified into six styles by considering their emotion arcs (i.e., the trajectory of average emotional tone expressed over the course of a story).

Based on this idea,~\citet{brahman-chaturvedi-2020-modeling} considered variations in the emotions expressed by story protagonists as time series to generate emotion-aware stories.
They split a five-sentence story into three segments, including a \textit{beginning} (the first sentence), a \textit{body} (the second to fourth sentences), and an \textit{ending} (the last sentence).
They applied labels for emotions in each segment to represent the emotional arc, and utilized them in story generation.

\subsection{Transformer-based Language Models for Vision and Language}
\label{subsec:transformer_for_v_and_l}

Transformer was proposed as an encoder-decoder model. Both the encoder and decoder parts include a self-attention mechanism.
Language models designed to be pre-trained using large unsupervised datasets have been proposed and have been shown to perform well in various natural language processing tasks.
Typical examples include Generative Pre-trained Transformer (GPT)~\cite{radford2018improving} and Bidirectional Encoder Representations from Transformers (BERT)~\cite{devlin-etal-2019-bert}.
Whereas these methods use only decoder and encoder parts, respectively, the Bidirectional and Auto-Regressive Transformer (BART)~\cite{lewis-etal-2020-bart} has been proposed as an encoder-decoder model with the intention of becoming more versatile model.

In recent years, the Transformer architecture has made a significant impact, not only on NLP, but also on other modalities. 
Moreover, multimodal applications such as those combining vision and language, have attracted the attention of researchers. 
There are two main approaches that combine vision and language processing in Transformer models, including two-stream and single-stream models.
Two-stream models~\cite{Lu_NEURIPS2019_c74d97b0, tan-bansal-2019-lxmert} utilize separate Transformers for each modality, and a cross-modality module is adopted.
Single-stream models~\cite{su2020vl-bert} directly input the text and visual embeddings into a single Transformer. 
\citet{su2020vl-bert} argued for the usefulness of using single-stream methods, as opposed to ViLBERT~\cite{Lu_NEURIPS2019_c74d97b0} and LXMERT~ \cite{tan-bansal-2019-lxmert}, each of which adopts a two-stream approach. 
They explained that the network architecture of the attention pattern in the cross-modal Transformer is restricted in ViLBERT and LXMERT, and proposed VL-BERT as a unified architecture based on Transformer models, without any restriction on the attention patterns, in which the visual and linguistic contents interact early and freely. 
Their LXMERT indicated that the single-stream method is promising.

The proposed method is based on UNITER (Universal Image-Text Representation)~\cite{chen2020uniter} as a single-stream model.
UNITER has shown excellent performance in many vision and language tasks, such as visual question answering (VQA) and image-text retrieval, with a relatively simple configuration of a single stream.

The method proposed in~\citet{Yu_2021_CVPR} is the closest approach to our work in the literature. 
The authors considered the Transitional Adaptation of a Pretrained Model (TAPM), which showed state-of-the-art performance on a visual storytelling task.
The main difference between the proposed approach and that of~\cite{Yu_2021_CVPR} is that we focus on image-narrative generation, whereas~\citet{Yu_2021_CVPR} applied their method to visual storytelling.
The task of visual storytelling was proposed by~\citet{huang2016visual}, along with an associated dataset called VIST. 
They explored visual storytelling as a sequential vision-to-language task. 
In contrast image narrative generation is a single vision-to-language task.
VIST has a one-to-one alignment between images and text.
In other words, visual storytelling generates a single sentence per image, while considering the preceding and subsequent images as well.
In contrast, the image narrative generation task, an entire narrative is generated from a single image.
We believe that image narrative generation is a more creative task, in which a narrative sequence must be imagined from the relatively limited data contained in a single image.
Moreover, to the best of our knowledge, the present work is the first to incorporate emotion arcs in image narrative generation to consider the importance of emotions in a story.

\subsubsection{Attention Mechanism in Transformers}

Transformer's attention mechanism was proposed based on the flow of introducing the attention mechanism into sequence-to-sequence model for machine translation \cite{Bahdanau-etal-2015-align-and-translate,luong-etal-2015-effective}.
Although the attention mechanism was introduced into the field of NLP as an auxiliary mechanism to compensate for information that cannot be captured by the recurrent neural network,~\citet{Vaswani_2017_transformer} proposed the Transformer as a model with attention as the main mechanism.

Transformer's attention mechanism maps a query and key-value pairs to an output. Their particular ``Scaled Dot-Product Attention'' can be described as follows.

\begin{equation}
    \rm{Attention}(Q, K, V) = \rm{softmax}\left( \frac{QK^T}{\sqrt{d_k}} \right)
\end{equation}

$Q$, $K$, $V$ stand for query, key, and value, respectively. $d_k$ is the dimension of $K$, and scaling factor $\frac{1}{\sqrt{d_k}}$ was introduced to prevent from dot product $QK^T$ growing large in magnitude and causing extremely small gradients in softmax function.

The encoder of the vanilla transformer contains self-attention layers. In a self-attention layer, queries and key-value pairs are come from the same place. 
The relationship, or attention, between each element of the sequence given as input and all other elements is calculated. 
In the decoder, each layer uses an output of the previous layer as queries, and key-value pairs come from the output of the encoder.

UNITER is a model that uses only the Encoder part of the Transformer, and embeds the input image and text, and then processes them with the self-attention mechanism.

Our method, ViNTER, is an encoder-decoder model like BART, and its encoder is a modified version based on UNITER.

\section{Proposed Method}
\label{sec:proposed_method}

We propose ViNTER (\textbf{Vi}sual \textbf{N}arrative \textbf{T}ransformer with \textbf{E}motion arc \textbf{R}epresentation) as a method to utilize emotion changes in narratives.
The overview of the model is shown in Figure \ref{fig:model}.
To encode an image and the desired emotions simultaneously, we developed ViNTER's encoder based on UNITER~\cite{chen2020uniter}, which is a well-known multimodal pre-trained Transformer model.
UNITER is a vision-and-language pre-trained model that can encode both visual and textual features with a single Transformer encoder.
After the image and emotion features are encoded, the Transformer decoder was trained to generate the corresponding narratives.
We describe each of the encoder and decoder in detail in the following sections.

\begin{figure}[!t]
\centering
\includegraphics[width=1.0\columnwidth]{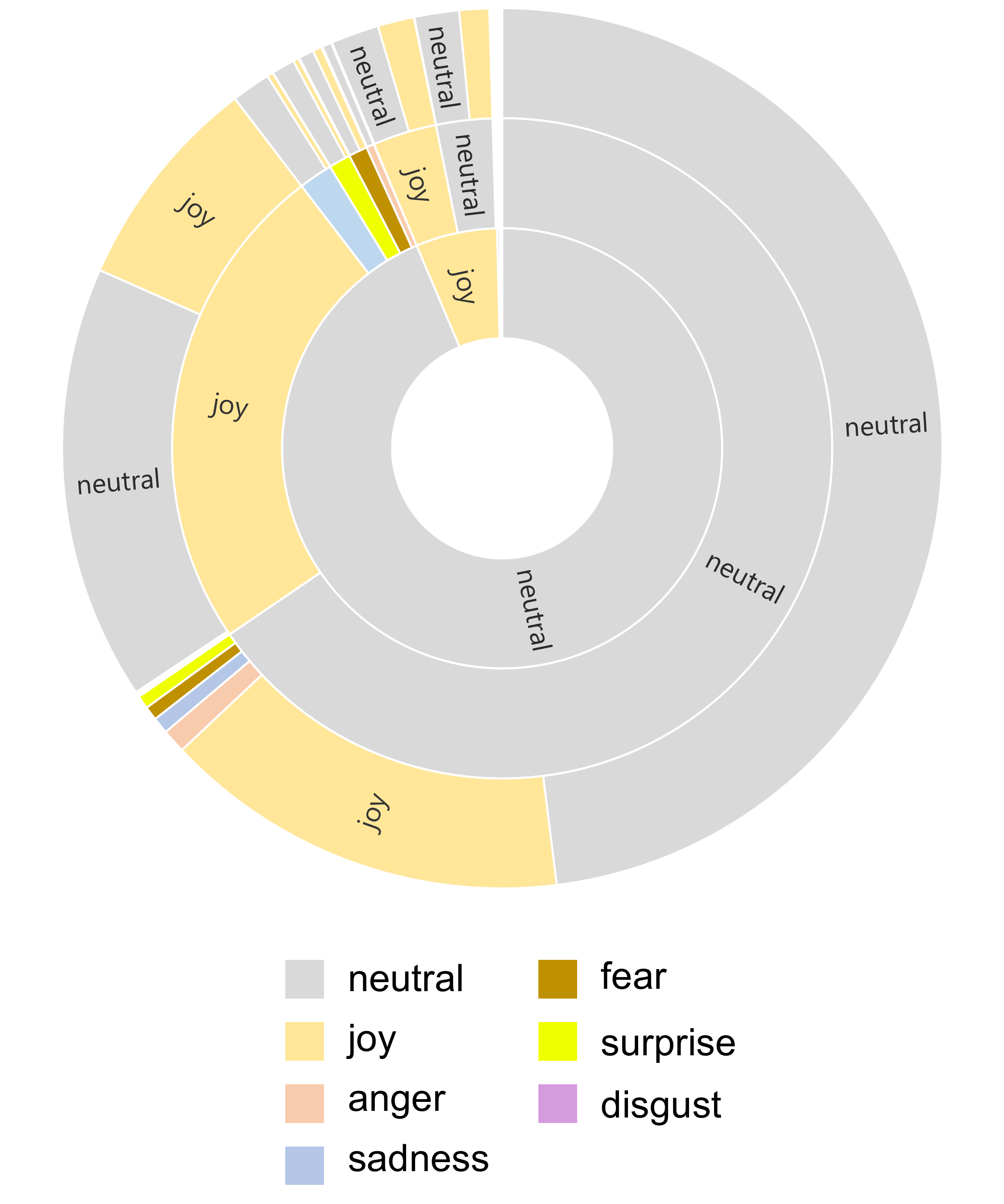}
\caption{
Visualization of the frequency of each emotion that appears in each segment of the emotion arc in the entire Image Narrative dataset.
From the center to the outside of the graph, the number of emotions contained in each segment (beginning, body, and end) is shown.
}
\label{fig:emotion_arcs}
\end{figure}

\subsection{Emotion Arc}\label{subsec:emotion_labeling}
The proposed approach adopts emotion arcs as an emotional representation of the input of the model.
Emotion arc represents the sequence of emotions expressed over the course of a story.
In this study, we consider a story that consists of five sentences.
We then split the story into three segments, including a \textit{beginning}, a \textit{body}, and an \textit{ending}, as in~\cite{brahman-chaturvedi-2020-modeling}.
We consider the sequences of emotions that match each segment as an emotion arc. 
For example, if the story begins with a joyful sentence, the body segment expresses fear, and the end segment expresses surprise, then the emotion arc is represented as ``joy $\rightarrow$ fear $\rightarrow$ surprise.''

To label sentences with emotion, we use an emotion classifier fine-tuned with GoEmotions.
GoEmotions is a corpus created to fine-tune emotion prediction models.\footnote{\url{https://github.com/google-research/google-research/tree/master/goemotions}}
Referring to~\cite{brahman-chaturvedi-2020-modeling}, we rely on Ekman's theory of basic emotions and use 6 + 1 categories (anger, disgust, fear, joy, sadness, surprise, and neutral).
We split the narratives into beginning, body, and end segments, and applied the emotion classifier to obtain the emotions corresponding to each segment.
A visualization of the number of each emotion that appeared for each segment of the emotion arc is shown in Figure~\ref{fig:emotion_arcs}.

\subsection{Encoder}\label{subsec:encoder}

\subsubsection{Visual Embeddings}

One of the inputs for the ViNTER encoder is image feature $\mathbf{I} \in \mathbb{R}^{n \times d}$.
$n$ indicates the number of object regions proposed by the object detector and $d$ indicates the dimensions of the image feature.
We use Faster R-CNN~\cite{faster_r_cnn} as an objection detection model, which is pre-trained with the Visual Genome dataset~\cite{visualgenome,butd}.
To explicitly provide positional information on the image region, we add positional embeddings based on the region coordinates, as used in UNITER. 
Specifically, we use the region coordinates represented as the 7-dimensional vector consisted of normalized top/left/bottom/right coordinates, width, height, and area.

\subsubsection{Emotion Embeddings}

In order to obtain the emotion labels of each segments, we use the BERT model fine-tuned with GoEmotions\footnote{\url{https://github.com/monologg/GoEmotions-pytorch}}.
As a taxonomy of emotions, we adopt seven emotions (anger, disgust, fear, joy, sadness, surprise, and neutral) based on Ekman's six basic emotions.

To input the emotion arc into the encoder, the proposed approach uses a special \texttt{[EMOTION SEP]} token to concatenate the words denoting each emotion (e.g., $\lbrack$ joy, \texttt{[EMOTION SEP]}, fear, \texttt{[EMOTION SEP]}, surprise$\rbrack$).
We used the learned word embeddings from BERT~\cite{devlin-etal-2019-bert} to convert the emotion-word tokens into emotion vectors $\mathbf{E} = \lbrace \bm{w}_{\textrm{begin}},\;\bm{w}_{\texttt{SEP}},\;\bm{w}_{\textrm{body}},\;\bm{w}_{\texttt{SEP}},\;\bm{w}_{\textrm{end}} \rbrace$.
We add the relative positional embeddings to indicate which word corresponds to which part of the story.

\subsubsection{Encoder Architecture}
Having obtained both visual and emotional embeddings, we concatenate them and input them into the Transformer encoder to obtain fused feature $\bm{h}$:
\begin{equation}
    \bm{h} = \textrm{Enc} ([\bm{I};\,\bm{E}])
\end{equation}
We use Transformer-based encoder to encode both of visual and emotional embeddings.
Each Transformer block consists of a stack of self-attention layers.

\subsection{Decoder}\label{subsec:decoder}

The decoder of ViNTER follows the decoder architecture of BART~\cite{lewis-etal-2020-bart}.
Our decoder is composed of a stack of Transformer blocks.
Each block includes multi-headed self-attention and cross-attention layers, and two fully-connected layers.
To avoid seeing the information of a future word, we apply a causal mask to the input of the self-attention layer.
Finally, we employ the fully-connected layer to compute the probability distribution over the entire vocabulary.
We trained our model parameters $\theta$ by minimizing the negative log-likelihood over the probability distribution of the ground-truth narrative $y$, as follows.
\begin{equation}
    \mathcal{L} = -\sum^{|y|}_{n=1}\log P_{\theta}(y_n\,|\,y_{<n},\;\mathbf{h}).
\end{equation}

\section{Experiment}
\label{sec:experiment}

\subsection{Dataset}
\label{subsec:dataset}

We used the \textbf{Image Narrative} dataset~\cite{shin2018customized} to conduct our experiments to validate the performance of the proposed approach.
This dataset contains 11,001 images drawn from MS COCO~\cite{lin2014microsoft} and 12,245 image narratives annotated through crowdsourcing.
Each narrative consists of five sentences.
The crowd workers were instructed to include local elements and sentiments that could be inferred or imagined from the images, not only accurately describe the contents of the images.
Because the test set of the Image Narrative dataset is not publicly available, we further split the training data of the original dataset into training and evaluation data, taking care to avoid overlapping images.
Consequently, we obtained a dataset with 5,398 narratives allocated to training and 1,813 to evaluation.

The narratives in this dataset contain noisy sentences; for example, some do not start with a capital letter or have extra or missing space around the period.
Therefore, we preprocess the sentences by removing or adding spaces around periods as needed, capitalizing the beginning of the sentences, and ensuring that the remainder was correctly spelled in lowercase.

\subsection{Implementation Details}
\label{subsec:implementation_details}

We set the image-feature dimension $d$ to 2,048 and the object region number $n$ to 36.
The number of hidden units of the Transformer block in the encoder and the decoder was set to 768.
The number of Transformer blocks in the encoder and decoder was set to 12.
We start our encoder training with the pre-trained checkpoints of UNITER\footnote{downloaded from https://github.com/ChenRocks/UNITER}.
We used the AdamW optimizer~\cite{loshchilov2018decoupled} with $(\beta_1 = 0.9,\;\beta_2 = 0.999)$.
We applied the cosine annealing schedule with an initial learning rate of $5.0\times10^{-5}$ and a warmup step of 1,000.
We trained the model for 5,000 steps with a batch size of 32.
The training took about two hours using a single Tesla A-100 GPU.

\begin{figure}[!t]
\centering
\includegraphics[width=1.0\linewidth]{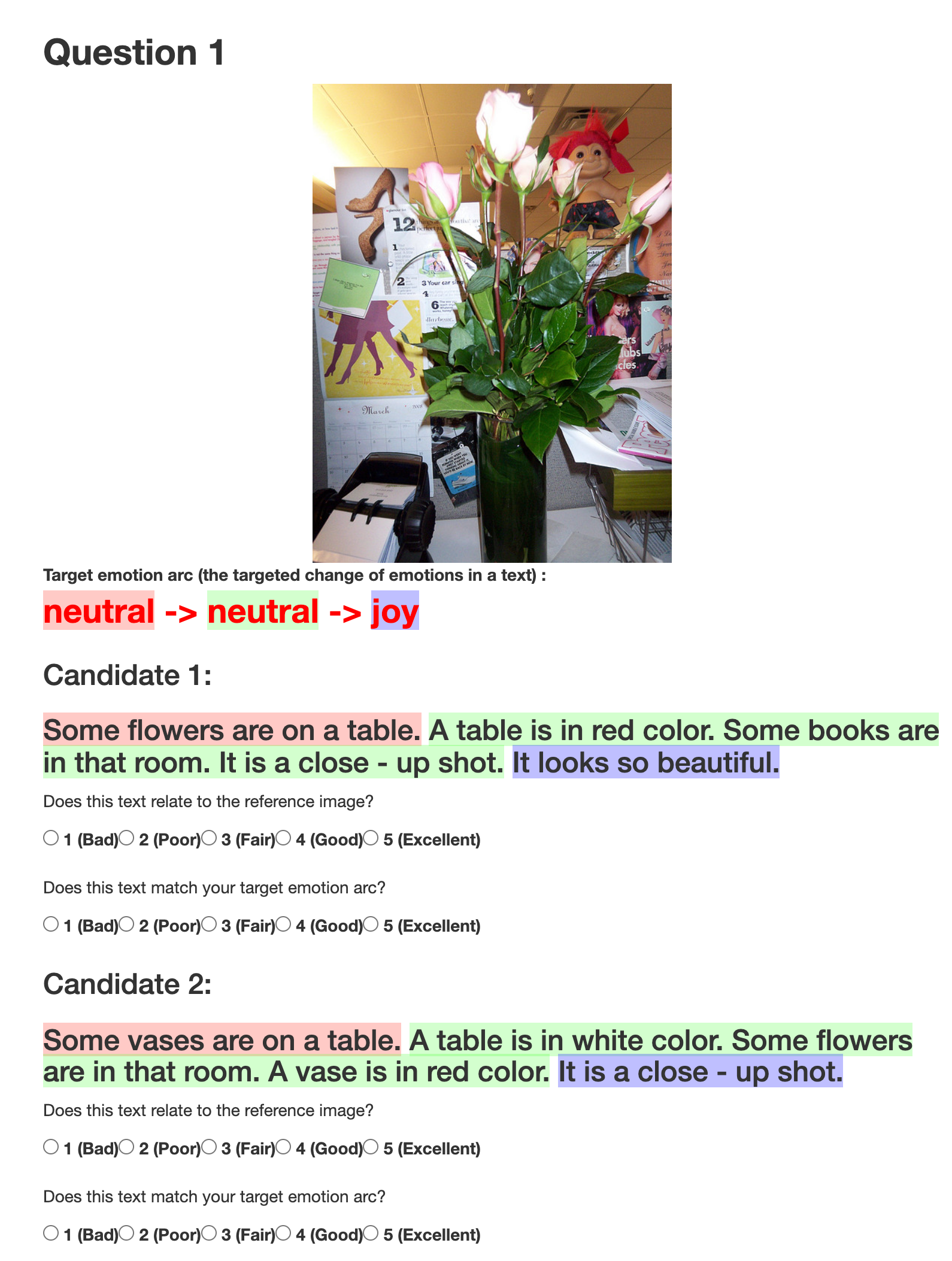}
\caption{
Screenshot of the evaluation interface presented to AMT workers.
The workers are asked to evaluate the quality of the narratives generated by ViNTER and the narratives generated by one of the comparison methods.
}
\label{fig:amt}
\end{figure}
\subsection{Baselines}\label{subsec:baselines}
We compared our method with the following baselines.

\textbf{DenseCap~\cite{Johnson_2016_CVPR}}:  This is a captioning model that generates captions focusing on a specific region.
Using DenseCap as a baseline is derived from the previous research~\cite{shin2018customized}.

\textbf{\citet{shin2018customized}}: This model generates narratives based on the user's interests through their answers to questions about the image.
For the above methods, we list the values as they appear in~\cite{shin2018customized}, which proposed the Image Narrative dataset.
Note that their results are evaluated on the test set of the dataset, which is not publicly available.
Hence, their scores are not directly comparable to ours; however, we include them for reference.

\textbf{Image Only}: This has the same model architecture as the proposed model, but used only image region features as input.

\textbf{ViNTER$_{\mathbf{begin,\;body,\;end}}$}: These methods are designed to control emotions in narratives in a simpler manner, using only a single emotion as input.
Each model uses an emotion corresponding to the beginning, body, and end segments as inputs.

\begin{table*}[t]
\centering
\begin{tabular}{@{}lcc@{}}
\toprule
Models                         & BLEU-4         & BERTScore      \\ \midrule
DenseCap~\cite{Johnson_2016_CVPR} ({\dag})                   & 1.90           & -              \\
\citet{shin2018customized} (\dag) & 1.41           & -               \\
Image Only                     & 7.288          & 88.40            \\
ViNTER$_{\mathrm{begin}}$      & 7.292          & \underline{88.46}        \\
ViNTER$_{\mathrm{body}}$       & \textbf{7.702} & 88.44        \\
ViNTER$_{\mathrm{end}}$        & 7.316          & \underline{88.46}   \\
\textbf{ViNTER}                & \underline{7.684}          & \textbf{88.50}  \\ \bottomrule
\end{tabular}
\caption{Automatic evaluation results in terms of the quality of the generated narratives.
For the models marked with ({\dag}), we put the values provided in~\cite{shin2018customized}.
Note that these models are evaluated on the test data of the dataset and cannot be directly compared to the results of the other models.
The best values are shown in bold and the second best values are underlined.
}
\label{tab:automatic_evaluation_metrics}
\end{table*}

\begin{table*}[t]
\centering
\begin{tabular}{@{}lccc|cc@{}}
\toprule
Models                    & Begin-acc      & Body-acc       & End-acc        & Seg-acc        & Arc-acc        \\ \midrule
Image Only                & 92.06          & 48.65          & 55.38          & 65.36          & 26.31          \\
ViNTER$_{\mathrm{begin}}$ & \textbf{93.16} & 51.85          & 53.83          & 66.28          & 27.25          \\
ViNTER$_{\mathrm{body}}$  & 91.23          & \textbf{66.46} & 55.43          & \underline{71.04}          & \underline{33.65}          \\
ViNTER$_{\mathrm{end}}$   & 91.73          & 52.73          & \textbf{61.89} & 68.78          & 30.72          \\
\textbf{ViNTER}           & \underline{93.05}          & \underline{65.97}          & \underline{60.89}          & \textbf{73.30} & \textbf{38.61} \\ \bottomrule
\end{tabular}
\caption{Automatic evaluation results in terms of emotional accuracy.
The best values are shown in bold and the second best values are underlined.
}
\label{tab:automatic_evaluation_metrics2}
\end{table*}

\subsection{Evaluation Metrics}
\label{subsec:evaluation_metrics}

\subsubsection{Automatic}
For evaluation by comparison with the ground-truth narratives, we use \textbf{BLEU}~\cite{papineni-etal-2002-bleu} and the F1 score of \textbf{BERTScore}~\cite{Zhang*2020BERTScore:}.
BLEU is a widely used metric in image captioning calculated based on n-gram matching between the generated and reference texts.
In recent years, research has been conducted on metrics that use text features obtained from a large-scale pre-trained language model.
BERTScore is a typical example of such a metric. 
It calculates the matching of tokens in generated and reference texts as conventional metrics, but it uses contextual embeddings calculated with BERT~\cite{devlin-etal-2019-bert} instead of string matching or matching heuristics.

We also evaluated the accuracy of the emotions expressed in the generated narratives.
We use 
\textbf{Begin-acc}, \textbf{Body-acc}, and \textbf{End-acc} as metrics to evaluate whether the emotions estimated for the generated narratives for the segments \textit{begin}, \textit{body}, and \textit{end}, respectively, are consistent with the corresponding parts of the input emotion arc.
\textbf{Seg-acc} is the fraction of matches between the correct emotion for each segment of the ground-truth and the estimated emotion for each segment of the generated narrative.
\textbf{Arc-acc} similarly measures the agreement of the estimated emotions; however, here we measure the agreement of the entire emotion arc, not each segment.

\subsubsection{Human Evaluation}
\label{subsubsec:human_evaluation}

We conducted the human evaluation with Amazon Mechanical Turk.\footnote{https://www.mturk.com/}
An example screenshot of the AMT task is shown in Figure~\ref{fig:amt}.

For comparison, we used four methods from the aforementioned baselines: Image Only, ViNTER$_{\mathrm{begin}}$, ViNTER$_{\mathrm{body}}$, and ViNTER$_{\mathrm{end}}$.
We paired the proposed method with each of the four comparison methods and conducted an evaluation experiment using 100 images.
13 workers were asked to evaluate a pair of image narratives generated by our proposed method and one of the comparison methods.

In order to compare how the change in emotion arc affects the methods, we also prepared the generated narratives in which the target emotion arcs were limited to those that changed in the middle of a narrative, i.e., always included two or more emotions. We distinguish these methods as ``w/ change,'' i.e. with change.
For each of the proposed and comparison methods, we also evaluated the generation of w/ change.

We asked following two questions to each worker for the evaluation:
\textit{(1) Does this text relate to the reference image?
(2) Does this text match the target emotion arc?}
These criteria are designed to evaluate whether the generated text is suitable as an image narrative and matches the required emotion arc. A five-point scale evaluation was used for each criterion: higher is better (5: Excellent, 4: Good, 3: Fair, 2: Poor, 1: Bad).

\section{Results and Discussion}\label{sec:results_and_discussion}

\begin{table*}[]
\centering
\begin{tabular}{@{}lcccc@{}}
\toprule
 & \multicolumn{4}{c}{\textbf{Better / Tie / Worse (\%)}} \\ %
 & Image & Emotion & Image (w/ change) & Emotion (w/ change) \\ \midrule
ViNTER vs. Image Only & \textbf{27.38} / 46.38 / 26.23 & 27.54 / 44.08 / \textbf{28.38} & \textbf{22.62} / 54.92 / 22.46 & \textbf{24.62}   / 55.38 / 20.00 \\
ViNTER vs. ViNTER$_{\mathrm{begin}}$ & 23.23 / 45.38 / \textbf{31.38} & \textbf{27.54} / 45.23 / 27.23 & \textbf{26.08} / 49.92 / 24.00 & \textbf{27.38}   / 46.31 / 26.31 \\
ViNTER vs. ViNTER$_{\mathrm{body}}$ & 25.23 / 47.62 / \textbf{27.15} & 27.23 / 45.54 / 27.23 & \textbf{31.69} / 39.85 / 28.46 & 26.54   / 42.15 / \textbf{31.31} \\
ViNTER vs. ViNTER$_{\mathrm{end}}$ & \textbf{28.38} / 44.46 / 27.15 & \textbf{27.08} / 47.54 / 25.38 & 29.15 / 41.46 / \textbf{29.38} & \textbf{33.77}   / 37.77 / 28.46 \\
\bottomrule
\end{tabular}
\caption{
Human evaluation results of the generated  narratives in terms of the relevance to the image and emotion arc.
Here, ``Better'' means the percentage of cases where the proposed method was evaluated as better quality than the baseline, ``Worse'' means the opposite, and ``Tie'' means the percentage of cases where the two methods were evaluated as equal.
We highlight the larger of the Better or Worse value in each set.
}
\label{tab:human-evaluation-result}
\end{table*}

\subsection{Automatic Evaluation}\label{subsec:automatic_evaluation_result}
The evaluation results of the automatic metrics in terms of the quality of the generated narratives are presented in Table~\ref{tab:automatic_evaluation_metrics}.

In BLEU and BERTScore, ViNTER outperforms the existing methods (ViNTER vs. DenseCap, Shin et al.).
This suggests that our method is superior to conventional methods, even though considering that these evaluations were performed using test data that cannot be used.
ViNTER also performs considerably better than the Image Only model.
However, when comparing the proposed method to models with a single emotion as input, in particular, ViNTER$_{\mathrm{body}}$, the difference in scores becomes smaller, or the comparison method scores slightly better.
This may be due to the shortage of metrics using comparisons with ground-truth data in the evaluation of narrative generation.
It is difficult to accurately evaluate the performance of models in reference-based metrics, when a variety of outputs are expected for the same input, which is exactly the case for narrative generation.

We also present the results in terms of the accuracy of the emotions of the generated narratives in Table~\ref{tab:automatic_evaluation_metrics2}.
The model with the emotion input for each segment has the best score for each corresponding segment (ViNTER$_{\mathrm{begin}}$ for Begin-acc, ViNTER$_{\mathrm{body}}$ for Body-acc, and  ViNTER$_{\mathrm{end}}$ for End-acc).
However, models that input emotions for a single segment have much lower accuracy compared to the other segments.
These results are in contrast to the results of ViNTER with an emotion arc as input, which scored the second-best for all segments.
In addition, ViNTER with an emotion arc input performs significantly better than the other methods in evaluating the accuracy of the emotions across the entire generated narrative (Seg-acc, Arc-acc).
These results show that our proposed method is capable of controlling emotional changes, not only in a part of the narrative but also in the complete narrative.

\subsection{Human Evaluation}\label{subsec:manual_evaluation_result}

The results of the human evaluation are shown in Table \ref{tab:human-evaluation-result}.
First, looking at the case of no change in emotion (left half of the table), ViNTER tends to be rated as having equal or slightly higher quality of emotional control compared to baseline.
In some cases, ViNTER is rated as worse in terms of the match between the image and the content of the generated narrative.
However, this is not particulary suprising given the fact that ViNTER is trained to generate narratives that follow not only the content of the image but also the emotion arc.
This suggests that ViNTER may have prioritized the emotion arc-based narrative generation over the image content in some cases.

In the case where there is a change in the emotion arc (right side of the Table.~\ref{tab:human-evaluation-result}), the superiority of ViNTER in terms of the quality of emotion is even more pronounced, except for vs. ViNTER$_{\textrm{body}}$.
This is an expected result, considering that when there is no change in the emotion arc, there is less likely to be a large difference in the generated narratives compared to the other methods.
Since the body part is the major part of the overall narrative, it is likely that humans would have tended to give a better evaluation of the overall emotion if the emotion of the body part had been generated correctly.
From these results, we can conclude that ViNTER is able to generate narratives that better capture the changes in emotion arc than the baseline method.

\begin{figure*}[t]
\centering
\includegraphics[width=0.7\linewidth]{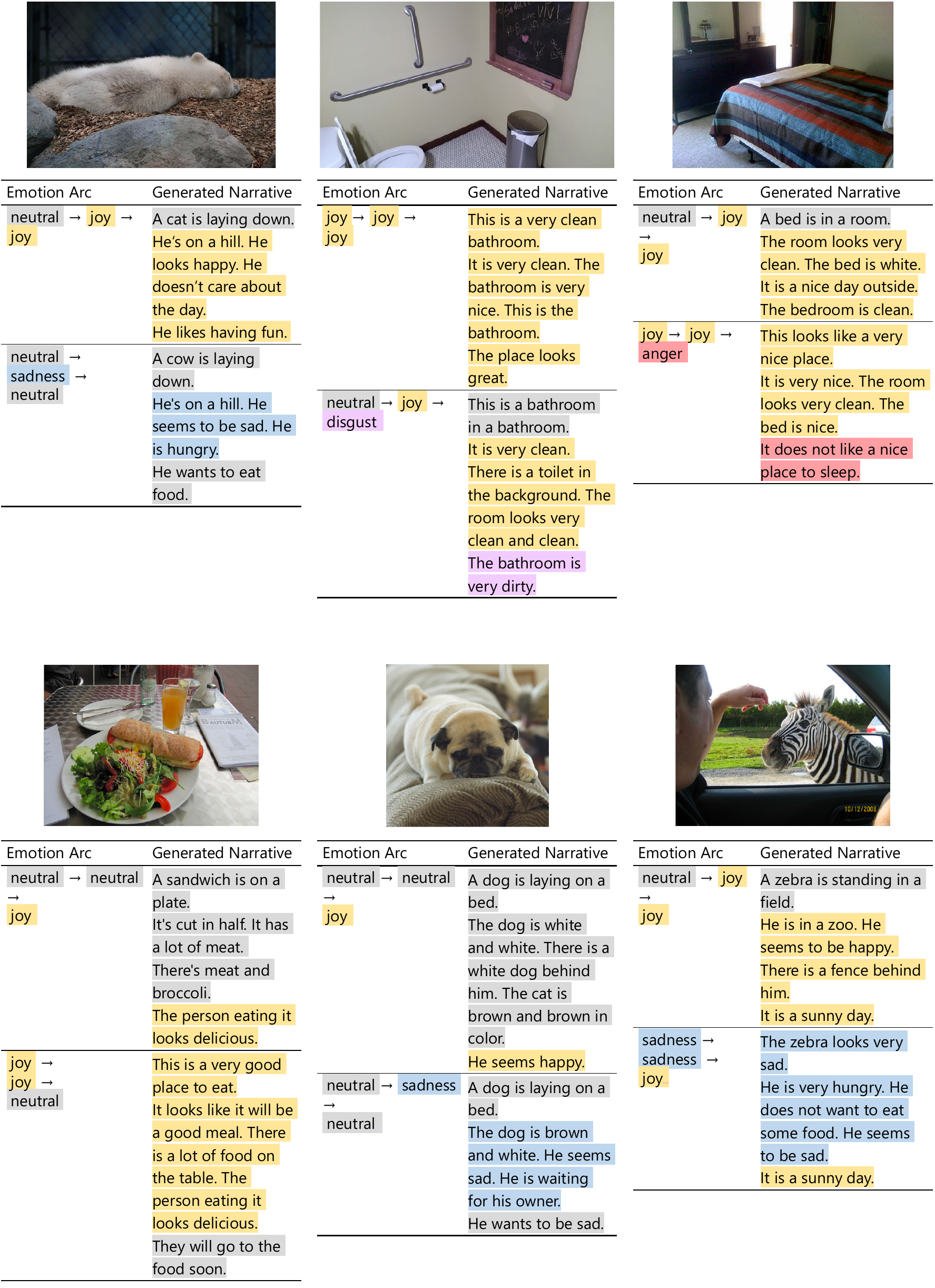}
\caption{
Qualitative examples for the same image with two different emotion arc inputs.
}
\label{fig:case_studies}
\end{figure*}

\subsection{Case Studies}\label{subsec:case_studies}

To demonstrate the controllability of the narratives generated by the proposed model, we show examples of the narratives generated from the same image with different emotion arcs in Figure~\ref{fig:case_studies}.
Overall, our ViNTER could generate different narratives for the same image, depending on the input emotion arc.
For example, in the top left image, when describing an animal, if ``joy'' is input as the middle emotion, a positive sentence, such as ``He looks happy,'' will be generated, whereas if ``sadness'' is input as the middle emotion, a negative sentence, such as ``He seems to be sad,'' will be generated.
However, in some cases, such as when ``joy'' is followed by a negative emotion (e.g., ``disgust'' or ``anger''), the output may be inconsistent with the previous sentence.

\section{Conclusion}

In this study, we have addressed the task of generating narrative sentences from an image using a Transformer-based method to process multimodal vision and language data.
Our proposed method has shown promising results on both automatic and human evaluation, and case studies showed that our method could control generation with an emotion arc.

However, some room for improvement remains in terms of the way these emotions are modeled.
The emotion arc only considers emotions from one perspective, but  multiple emotions are typically associated with a given narrative. 
\citet{Karl_2005_writing_for_emotional_impact} insisted on the importance of distinguishing character and reader emotions.
Although the handling of emotions in this study was simplistic, we believe that this work is an important step toward utilizing more diverse and difficult-to-handle emotions in story generation.

In parallel with creating a more sophisticated method, it is also essential to build a dataset to train and measure the performance of that method.
A larger dataset that includes more diverse range of emotions with less bias is desirable.

For further discussion on the importance of ``creativity'' in image narrative generation for the social implementation of creative AI, we refer to the idea of creative systems.
\citet{Colton2008CreativityVT} introduced the notion of the creative tripod.
The three legs represent the three behaviors required in their system, including skill, appreciation, and imagination.
Later, \citet{elgammal2017can} rephrased this notion as the ability to produce novel artifacts (imagination), the ability to generate high-quality artifacts (skill), and the ability of a model to assess its own creations.
Among these three goals of creative systems, our proposed method has achieved ``imagination'' and ``skill.''
However, the ability to assess creativity has not been established.
To achieve this, a system must be developed that can evaluate the image narratives it generates by itself; that is, the system must be able to learn criteria for a good image narrative.
As may be noted from the discussions of the results of the human evaluation, the assessment of creativity is not an easy task, even for humans.
We expect the development of methods to automatically evaluate narratives would be of considerable benefit in enabling great strides towards the realization of creative AI.

\section*{Acknowledgements}
This work was partially supported by JST AIP Acceleration Research JPMJCR20U3, Moonshot R\&D Grant Number JPMJPS2011, JSPS KAKENHI Grant Number JP19H01115, and JP20H05556 and Basic Research Grant (Super AI) of Institute for AI and Beyond of the University of Tokyo.
We would like to thank Thomas Westfechtel, Kenzo Lobos-Tsunekawa, and Yuki Kawana for the helpful discussions.

\bibliographystyle{ACM-Reference-Format}
\bibliography{cameraready}

\end{document}